\documentclass{article}

\usepackage{PRIMEarxiv}

\usepackage[utf8]{inputenc} 
\usepackage[T1]{fontenc}    
\usepackage{hyperref}       
\usepackage{url}            
\usepackage{booktabs}       
\usepackage{amsfonts}       
\usepackage{nicefrac}       
\usepackage{microtype}      
\usepackage{lipsum}
\usepackage{fancyhdr}       
\usepackage{graphicx}       
\graphicspath{{media/}}     

\usepackage{setspace}

\usepackage{enumitem}
\usepackage{amsmath}
\usepackage{amssymb}

\usepackage{tabularx}
\usepackage{multirow}

\usepackage{algorithm}
\usepackage{algorithmic}

\title{Advanced Weakly-Supervised Formula Exploration for \\ Neuro-Symbolic Mathematical Reasoning}

\author{
  \textbf{Yuxuan Wu} and \textbf{Hideki Nakayama} \\
  The University of Tokyo \\
  \texttt{\{wuyuxuan, nakayama\}nlab@ci.i.u-tokyo.ac.jp} \\
}

\begin{document}
\maketitle

\setstretch{1.1}

\begin{abstract}
In recent years, neuro-symbolic methods have become a popular and powerful approach that augments artificial intelligence systems with the capability to perform abstract, logical, and quantitative deductions with enhanced precision and controllability. 
Recent studies successfully performed symbolic reasoning by leveraging various machine learning models to explicitly or implicitly predict intermediate labels that provide symbolic instructions. 
However, these intermediate labels are not always prepared for every task as a part of training data, and pre-trained models, represented by Large Language Models (LLMs), 
also do not consistently generate valid symbolic instructions with their intrinsic knowledge.
On the other hand, existing work developed alternative learning techniques that allow the learning system to autonomously uncover optimal symbolic instructions.
Nevertheless, their performance also exhibits limitations when faced with relatively huge search spaces or more challenging reasoning problems.
In view of this, in this work, we put forward an advanced practice for neuro-symbolic reasoning systems to explore the intermediate labels with weak supervision from problem inputs and final outputs. Our experiments on the Mathematics dataset illustrated the effectiveness of our proposals from multiple aspects.


\end{abstract}

\section{Introduction}

In recent years, rapid progress in deep learning has yielded remarkable achievements across various areas and tasks from computer vision to natural language processing.
Nevertheless, the solutions to mathematical problems are still often considered a formidable challenge due to the demand for precise logical and quantitative calculation and deduction,
which is discordant with the fundamental nature of deep learning models
that rely on stacked layers of neurons for estimations and fittings.
To address these constraints of neural networks, researchers gravitated towards neuro-symbolic methods.
The concept of neuro-symbolic learning is vast, and its origin dates back to the last century \cite{ns0}.
Modern neuro-symbolic practices also have integrated symbolic assistance at various learning stages adapting to the distinct characteristics and requirements of varied tasks \cite{ns1,ns2}.
Specific to mathematical reasoning tasks, neuro-symbolic approaches typically comprise symbolic practice by employing neural networks to receive and interpret question texts
and generate symbolic notations as instructions. 
Subsequently, these symbolic notations are deciphered with a calculator or interpreter in a rule-based manner to acquire the final answer.
This approach empowers the learning system to leverage the generalization capacity and end-to-end training of neural networks along with the precision and controllability of symbolic calculations.
This advantageous feature allows neuro-symbolic methods to achieve impressive performance across various mathematical reasoning tasks and datasets \cite{math23,mathqa,gsm8k}.

However, a pivotal concern underlying the success of neuro-symbolic methods is associated with the training of their inference models.
From the perspective where mathematical reasoning tasks are viewed holistically, the final answers, as the destination of reasoning, are naturally regarded as the inherent labels in these tasks.
In contrast, the symbolic instructions are more akin to intermediate labels and thus may not always be prepared for every task as a part of training data,
although they are indispensable to the training of inference models.
Confronted with this issue, early studies in this field manually annotated these symbolic labels to facilitate the application and validation of neuro-symbolic practices \cite{mawps,math23,mathqa}.
However, such annotations can be laborious, costly, and confined to specific datasets.
Recent work devised a novel approach that employs LLMs to generate these symbolic instructions and incorporates calculators to execute precise calculations \cite{gsm8k,cot}.
Nevertheless, there is no theoretical assurance confirming that LLMs can consistently produce valid instructions based on their intrinsic knowledge,
and those invalid symbolic instructions can subvert the overall efficacy of neuro-symbolic approaches.

On the other hand,
some subsequent studies introduced weakly-supervised learning methods that allow the autonomous discovery of intermediate symbolic instructions with end-to-end weak supervision \cite{math_fix,coling}. 
These methods carried out a preliminary capability to implement neuro-symbolic practices in scenarios without prepared symbolic annotations.
However, their performance also exhibits limitations when faced with relatively huge search spaces or more challenging reasoning problems.
Moreover, given that most existing studies on neuro-symbolic mathematical reasoning have focused on formulating tree-structured computations, these weakly-supervised learning proposals also may not effectively and efficiently address the exploration of symbolic instructions in more general forms that support non-tree-structured computations.
All these factors limited the application of existing weakly-supervised neuro-symbolic learning approaches in more general learning scenarios.

In view of this, 
in this work, we are motivated to develop an advanced learning framework for neuro-symbolic approaches in mathematical reasoning problems with enhanced capability to explore intermediate symbolic labels with weak supervision from problem inputs and final outputs.
With these symbolic labels referred to as formulas, our general idea aligns with the principle of existing work that implements a recurrent exploration process in which the optimal formula for each problem is gradually uncovered and the inference model is continuously optimized.
Nevertheless, we put forward multiple fresh proposals and improvements to the existing work to further enhance the formula exploration efficiency.
At the same time, we are also motivated to integrate this learning framework with more general formula representation techniques to enhance the expressive capacity of symbolic solutions.
On the basis of prior studies, our major contributions in this work encompass that:

\begin{itemize}[leftmargin=20pt]
    \item We extend the application of neuro-symbolic approaches with weakly-supervised learning in mathematical reasoning tasks to a more generalized symbolic notation system that supports functional Domain Specific Languages.
    \item We enhance the efficiency of the formula exploration technique proposed in existing work from multiple aspects to extend its applicability in more challenging and general mathematical problems and datasets.
    \item We demonstrate the feasibility of neuro-symbolic approaches powered by our learning technique within a fully weakly-supervised learning scenario devoid of formula annotations, as well as its superiority in comparison to end-to-end and LLM-based approaches on the Mathematics dataset \cite{math_dm}.
\end{itemize}

\section{Related Work}

\subsection{Mathematical Reasoning Tasks}

Mathematical reasoning tasks comprise a series of tasks developed to examine the ability of machine learning systems to solve mathematical problems and conduct abstract, logical, and quantitative reasoning.
Early studies in this field started with collecting and solving applied mathematical problems described in natural languages, which are also known as math word problems.
Several representative datasets for math word problems include MAWPS \cite{mawps}, Math23K \cite{math23}, and MathQA \cite{mathqa}.
Pushing the boundaries beyond applied mathematical problems, 
Saxton et al. constructed a large-scale dataset offering mathematical problems generated in a wide spectrum of mathematical areas \cite{math_dm},
and other work has also introduced datasets specialized to geometric problems \cite{math_g0} and theorem proving \cite{math_atp,ldj}.
Upon entering the era of LLMs, recent work published datasets collating mathematical problems together with their solutions annotated in natural languages \cite{gsm8k}.
Such tasks have also become an important benchmark in the evaluation of LLMs \cite{mmlu}.

\subsection{Mathematical Problem Solvers}

While it is feasible to train machine learning models to solve mathematical problems through end-to-end predictions, 
existing studies have revealed significant limitations in their generalization capacity,
which is generally because end-to-end inference inevitably receives and predicts rational numbers by digits or subwords, and thus results in the absence of the concept of numbers in their entirety \cite{math_dm}.
In contrast, an alternative approach that has been broadly adopted contemporarily is to make use of intermediate labels named equations or formulas \cite{mawps,math23,mathqa}.
This approach carried out a neuro-symbolic paradigm in which these intermediate labels, usually predicted by neural networks, can be interpreted as specific computations and executed by external tools to acquire problem answers.
Most of the research adopting this approach requires these intermediate labels to be prepared in the training data to permit straightforward training on the inference model.
Some other studies have also developed methods to learn these labels utilizing weak supervision derived from questions and answers \cite{math_fix,coling}.

On the other hand, with the recent advancements in large language models (LLMs), recent studies have also investigated the feasibility of prompting LLMs to solve mathematical problems. 
This approach is also known as Chain-of-Thought (CoT) because of LLMs' proficiency in generating the solution to problems as a chain that formulates the sequential reasoning steps \cite{palm,cot}.
This practice can be realized either through fine-tuning pretrained language models or utilizing few-shot or even zero-shot prompting \cite{zs1,zs2,zsr}.
Being aware of the precision issue exposed in mathematical calculations performed by LLMs, researchers also integrated CoT with neuro-symbolic computations by implementing CoT to generate the expressions for calculations and delegating the computations to external tools \cite{gsm8k}.
Similarly, Wolfram developed a solution equipping GPTs with the APIs of Wolfram, which merged the comprehensive reasoning capability of LLMs and the computational prowess of Wolfram \cite{wolfram}.

\section{Problem Definition}

\setlength{\tabcolsep}{6pt}
\begin{table}[t]
\centering
\caption{An example of a typical mathematical reasoning problem.} \label{tab:1}
\begin{tabularx}{0.9\textwidth}{r X}
\toprule
question & \texttt{Let a be 6 * 2.5 - 11, Find the value of b so that a ** 2 + b = 30.}\\
answer & \texttt{14} \\
\midrule
$q$ & \texttt{Let a be $\langle$N0$\rangle$ * $\langle$N1$\rangle$ - $\langle$N2$\rangle$, Find the value of b so that a ** $\langle$N3$\rangle$ + b = $\langle$N4$\rangle$ .}\\
$num$ & \texttt{(6, 2.5, 11, 2, 30)} \\
$a$ & \texttt{14} \\
$f$ & \texttt{[($\times$, N0, N1), (-, M0, N2), (\^{}, M1, N3), (-, N4, M2)]} \\
\bottomrule
\end{tabularx}
\end{table}

In this work, we study the practice of neuro-symbolic methods and verify our proposals on the mathematical reasoning task specific to question-answering problems. 
In this task, the basic unit of learning and evaluation is the pair of questions and answers.
An example is presented in Table \ref{tab:1}.
As for the preprocessing, we extract the numerical tokens present in each question and substitute these tokens with token \texttt{$\langle$Ni$\rangle$}, where \texttt{i} indicates the index of the original token within the question text. 
We let $q$ denote the question text with replaced tokens, and $num$ denote the tuple of extracted numbers.
We also refer to $q$ as the ``template'' because multiple raw questions can be converted to the same $q$ if the differences in their question texts are confined to the numerical tokens.
In this case, these questions are consolidated to a single $q$, and their corresponding $num$ and $a$ are collected into sets \{$num$\} and \{$a$\}.

On the other hand, we let $f$ denote the symbolic instructions expected to be produced by the learning system for solving the corresponding question.
They are also referred to as ``formula'' in the following discussion.
Theoretically, $f$ can be arranged in any format as long as it conveys unambiguous semantic meaning and can be executed by an associated interpreter. 
In this work, we implement $f$ with a sequential and functional Domain Specific Language (DSL).
A formal definition of this DSL is provided by Equations \ref{equ:f1} to \ref{equ:f4}.
As shown by the example in Table \ref{tab:1}, a formula $f$ comprises a sequence of basic terms [$S_{0}$, ..., $S_{n}$]. 
Each of these terms $S_{i}$ can be regarded as a function call, which comprises an operator $opr$ and a certain number of operands $opd_1$ to $opd_k$.
Given an interpreter and specific $num$, 
these terms can be computed in order from $S_{0}$ to $S_{n}$, and the computation result of the last term $S_{n}$ yields the result of the entire formula.
With $\mathbb{I}(\cdot)$ denotes a formula interpreter, we have $\hat{a} = \mathbb{I}(f,\ num)$.

\vspace{-10pt}
\begin{align}
    f &:= [S_{0}, S_{2}, ..., S_{n}] \label{equ:f1}\\
    S_{i} &:= (opr, opd_1, ..., opd_k) \\
    \{opr\} &:= \{+,\ -,\ \times,\ \div,\ \string^\ ,\ \texttt{ABS}, ...\}
\end{align}
\begin{align}
    \{opd\} &:= T_{num} \cup T_{re\hspace{-0.08em}f} \cup T_{con} \nonumber\\
    &:= \{\texttt{N0},\,\texttt{N1},\,...\,|\,\texttt{M0},\,\texttt{M1},\,...\,|\,1,\,2,\,\pi,\,...\} \label{equ:f4} 
\end{align}

More specifically, with regard to the composition of $opr$ and $opd$ in mathematical reasoning tasks, \{$opr$\} typically encompasses basic algebraic operations, common mathematical functions such as rounding and finding absolute values, and some specific functions tailored to the requirements of the areas that questions cover.
On the other hand, \{$opd$\} comprises operands available from the union of three sets: $T_{num}$, $T_{re\hspace{-0.08em}f}$, and $T_{con}$. 
Here, $T_{num}$ denotes the set of number tokens, of which each token \texttt{Ni} will be substituted with the $i$-th extracted number in $num$ during the computation of $f$. 
$T_{re\hspace{-0.08em}f}$ denotes the set of reference tokens, of which each token \texttt{Mj} will be substituted with the computation result of the previous $j$-th term $S_{j}$ during the computation.
$T_{con}$ denotes the set of constant tokens referring to the constant values that may not explicitly appear in question texts and $num$ but are indispensable for solving particular problems. 
They will also be substituted with the corresponding constant values during the computation.

Given the data preprocessing and formalization presented above, our goal in this task can be defined as follows. 
For each independent question template $q$, we regard the triplet ($q$,\ \{$num$\},\ \{$a$\}) as the basic instance for the learning process. 
For each instance of ($q$,\ \{$num$\},\ \{$a$\}), the objective is to guide the learning system to explore an optimal formula $f$ 
so that $\{\hat{a}\} = \mathbb{I}(f,\ \{num\})$ yields highest matching rate between $\{\hat{a}\}$ and $\{a\}$.

\section{Proposed Method}

\subsection{General Learning Procedure}

\begin{figure}[b]
\centering
\begin{minipage}{0.9\textwidth}
\begin{algorithm}[H]
\caption{General Learning Procedure} \label{alg:1}
\begin{algorithmic}[1]
\STATE \texttt{PolicyNet}\ $\gets$ \texttt{Init}()
\STATE $\mathbb{L}$ $\gets$ \{\}
\STATE \textbf{for} $loop$ \textbf{in} range($max\_loop$) \textbf{do}
    \STATE \hspace{1em} ($q$,\ \{$num$\},\ \{$a$\}) $\gets$ \texttt{Sample}($\mathbb{D}$)
    \STATE \hspace{1em} $f$, $accu_f$ $\gets$ \texttt{Search}(\texttt{PolicyNet},\ $q$,\ \{$num$\},\ \{$a$\}) \label{line:alg_search}
    \STATE \hspace{1em} $\mathbb{L}$.update($q$: ($f$,\ $accu_f$))
    \STATE \hspace{1em} \texttt{PolicyNet}.train($\mathbb{L}$) \label{line:train}
    \STATE \hspace{1em} \texttt{Reflect}(\texttt{PolicyNet}, $\mathbb{L}$) \label{line:reflect}
\STATE \textbf{end for}
\end{algorithmic}
\end{algorithm}
\end{minipage}
\end{figure}

The general learning procedure adopted in our study is formulated with Algorithm \ref{alg:1}.
Its fundamental idea is to implement a recurrent exploration process in which optimal $f$ for each $q$ is gradually acquired and the inference model is continuously optimized.
Here, \texttt{PolicyNet} denotes the inference model that takes $q$ as input and predicts the formula $f$.
$\mathbb{D}$ denotes the preprocessed dataset containing learning instances \{($q$,\ \{$num$\},\ \{$a$\})$_i$\}.
$\mathbb{L}$ denotes a dictionary recording the optimal formula $f$ together with the accuracy it achieved for each $q$. $\mathbb{L}$ is initialized as empty.

After initialization, the learning procedure is organized into numerous basic loops.
Within each loop, we first employ curriculum learning to determine the instance ($q$,\ \{$num$\},\ \{$a$\}) for the learning in the current loop with a \texttt{Sample} function, of which the details are presented in Section \ref{sec:sample}.
Then, we conduct a \texttt{Search} procedure with the purpose of searching for an optimal formula $f$ for the sampled instance, of which the details are presented in Section \ref{sec:search}.
After the formula $f$ and its corresponding accuracy $accu_f$ is returned by \texttt{Search}, 
$\mathbb{L}$ is updated with $f$ and $accu_f$.
Concretely, if $f$ is not \texttt{None}, and if either no $f$ has been recorded for $q$ in $\mathbb{L}$ or $accu_f$ surpassed the previously recorded accuracy, $f$ and $accu_f$ will be updated for $q$ in $\mathbb{L}$.
After this, \texttt{PolicyNet} is trained with the ($q$, $f$) pairs recorded in $\mathbb{L}$.
At the end of each learning loop, we introduced a reflection mechanism to reduce the redundancy of formulas and try rectifying formulas whose effectiveness is misjudged. 
The details of this \texttt{Relect} procedure are presented in Section \ref{sec:clean_reflect}.

\subsection{Problem Sampling} \label{sec:sample}

Given the fact that the difficulty of mathematical problems within a dataset may vary largely, and the solutions to relatively difficult problems usually lead to extremely huge spaces of formulas that cannot be well-explored in a reasonable time frame without prior knowledge, we bring in the idea of curriculum learning to refine the random sampling adopted by existing work.
Ideally, we encourage the learning to begin with easy samples and gradually transition to difficult ones. 
To achieve this, we empirically assess the difficulty of problems with the length of question texts (i.e., the number of tokens in $q$, namely \texttt{len}($q$)), and maintain the dataset $\mathbb{D}$ in three subsets: \{$new$\}, \{$unsolved$\}, and \{$solved$\}.
Among them:

\begin{itemize}[leftmargin=20pt]
    \item \{$new$\} comprises learning instances that have never been encountered, and it organizes these instances in an ordered list sorted by \texttt{len}($q$) in ascending order.
    When learning samples are requested from \{$new$\}, it consistently returns the first instance, which is the one with the minimal \texttt{len}($q$).
    \item \{$unsolved$\} comprises learning instances that have been previously sampled and subjected to the formula search. However, the formula search failed to acquire a formula that satisfied the acceptable threshold $\gamma$ for question answering accuracy. In this case, these instances are stored in \{$unsolved$\} and can be sampled again in future learning iterations.
    \{$unsolved$\} organizes these instances in an unordered set. When learning samples are requested from \{$unsolved$\}, it returns a random instance.
    \item \{$solved$\} archives learning instances on which the accuracy exceeds the acceptable threshold. These instances will not be sampled again.
\end{itemize}

At the beginning of the learning, all learning instances are held by set \{$new$\}, and sets \{$unsolved$\} and \{$solved$\} are empty.
As the learning progresses, the learning instances are distributed among these three sets, and the \texttt{Sample} procedure acquires new learning instances from set \{$new$\} and \{$unsolved$\} with probability $p$ and $(1-p)$, respectively.
With $N_{us}$ and $N_{s}$ denote the counts of learning instances currently held by \{$unsolved$\} and \{$solved$\}, respectively, 
we have $p$ $=$ $max(e^{-\frac{N_{us}}{N_{s}+1}}, 0.5)$.
This empirical setting encourages the learning procedure to revisit unsolved problems again when there are way more unsolved problems than solved ones.
This prevents the learning procedure from encountering difficult problems prematurely.
Nevertheless, it preserves a fifty percent possibility of fetching new learning instances.

\subsection{Formula Search} \label{sec:search}

As for the \texttt{Search} procedure executed in line \ref{line:alg_search} of Algorithm \ref{alg:1}, its detail is presented by Algorithm \ref{alg:2}.
For this procedure, we basically followed and improved the search algorithm proposed by prior study \cite{coling}.
The general idea of this algorithm is to organize a heuristic search on a graph structure with consideration of the confidence level of each formula and the formulas being close.

\begin{figure}[b]
\centering
\begin{minipage}{0.9\textwidth}
\begin{algorithm}[H]
\caption{Formula Search} \label{alg:2}
\begin{algorithmic}[1]
\STATE \textbf{func} \texttt{Search}(\texttt{PolicyNet},\ $q$,\ \{$num$\},\ \{$a$\})
    \STATE \hspace{1em} $\mathcal{G}$ $\gets$ \texttt{Init}() \label{line:init}
    \STATE \hspace{1em} \textbf{for} $iter$ \textbf{in} range($max\_iter$) \textbf{do}
        \STATE \hspace{2em} $f_{exp}$ $\gets$ \texttt{Sample}($\mathcal{G}$) \label{line:sample}
        \STATE \hspace{2em} \{$f_{mut}$\} $\gets$ \texttt{Mutate}($f_{exp}$) \label{line:mutate}
        \STATE \hspace{2em} $\mathcal{G}$.update(\{$f_{mut}$\}) 
    \STATE \hspace{1em} \textbf{end for}
    \STATE \hspace{1em} $f_{best}$ $\gets$ $\mathop{\arg\max}_{f \in \mathcal{G}}$ $f.accu$
    \STATE \hspace{1em} $f_{best}$ $\gets$ \texttt{CleanUp}($f_{best}$) \label{line:cleanup}
    \STATE \hspace{1em} \textbf{return} $f_{best}$, $f_{best}.accu$
\end{algorithmic}
\end{algorithm}
\end{minipage}
\end{figure}

\subsubsection{Formula Graph}

In Algorithm \ref{alg:2}, $\mathcal{G}$ denotes a graph structure utilized to store the formulas under exploration in each \texttt{Search} attempt. $\mathcal{G}$ is maintained under the following two rules:

\begin{itemize}[leftmargin=20pt]
    \item Each of its nodes $n_f$ represents a unique formula $f$.
    \item There is an edge between two nodes if and only if the distance between the formulas they represent is one.
\end{itemize}

Here, we measure the distance between formulas in terms of formula term $S_i$.
That is, if a formula $f$ can be modified to $f'$ by inserting, deleting, or modifying one single term, we consider the distance between $f$ and $f'$ as one.
Moreover, we let each node $n_f$ maintain a score $n_f.score$ that indicates the confidence level of each formula on solving the given $q$.
This score is calculated with Equation \ref{equ:s1} to \ref{equ:s3}.
Generally, this score is composed of two factors, ${p}(f|q,\theta_{P})$ and $f.accu$, which indicate the confidence level of $f$ judged by the \texttt{PolicyNet} and actual examination, respectively.
Here, $\beta$ is a hyperparameter balancing the influence of these two factors.
With $\theta_{P}$ denotes the parameters of \texttt{PolicyNet}, ${p}(f|q,\theta_{P})$ denotes the normalized likelihood of $f$ on $q$ given by \texttt{PolicyNet} following Equation \ref{equ:s2}.
In this calculation, we take the arithmetic mean of the likelihood on each formula term $S_i$ to mitigate the influence of variable formula length.
On the other hand, we measure $f.accu$ simply with the matching rate between the estimated answer s$\{\hat{a}\} = \mathbb{I}(f,\ \{num\})$ and the ground-truth answers $\{a\}$ following Equation \ref{equ:s3},
$\textbf{1}_{|\{\hat a\}_{i} - \{a\}_{i}| < 10^{-3}}$ returns one if the difference between the $i^{\textrm{th}}$ result in $\{\hat a\}$ and the $i^{\textrm{th}}$ answer in $\{a\}$ is less than an acceptable floating-point error bound $10^{-3}$.
Note that the calculation by $\mathbb{I}$ may not necessarily succeed because arithmetic errors, such as division by zero, can be encountered. 
In such cases, corresponding $\{\hat a\}_{i}$ is considered invalid and $|\{\hat a\}_{i} - \{a\}_{i}|$ is considered infinite.

\begin{align}
    n_f.score &= {p}(f|q,\theta_{P})+\beta*f.accu\, \label{equ:s1}\\
    {p}(f|q,\theta_{P}) &= \ \prod_{i=1}^{L}p(S_i|S_{1:i-1},q,\theta_{P}) \label{equ:s2}\\
    f.accu &= \ \frac{1}{N} \sum_{i=1}^{N}\textbf{1}_{|\{\hat a\}_{i} - \{a\}_{i}| < 10^{-3}} \label{equ:s3}
\end{align}

\subsubsection{Graph Initialization}

As shown in line \ref{line:init} of Algorithm \ref{alg:2}, 
the graph $\mathcal{G}$ is initialized at the beginning of each \texttt{Search} attempt.
During initialization, nodes representing formulas of the following three types are added to $\mathcal{G}$:

\begin{itemize}
    \item \texttt{[(+, N0, N1)]}, which is a universal starting point of search by default.
    \item $N$ formulas predicted by \texttt{PolicyNet} given $q$ with top-$N$ likelihood.
    \item $M$ formulas recorded in $\mathbb{L}$ for the $M$ templates that are semantically closest to the current given $q$.
\end{itemize}

Following this configuration,
$\mathcal{G}$ should be initialized with a maximum of $(N\!+\!M\!+\!1)$ initial nodes, while some of them can be merged if there are overlaps among these three types of initial nodes and duplications in the $(N\!+\!M\!+\!1)$ formulas.
Here, we determine the semantic distance between two templates $q$ and $q'$ by calculating the Euclidean distance between their sentence embedding generated by \texttt{PolicyNet}. That is, with $E_{P}(\cdot)$ denotes the encoder of \texttt{PolicyNet}, $distance(q, q') = \|E_{P}(q) - E_{P}(q')\|_2$.

\subsubsection{Formula Sampling}

As shown in line \ref{line:sample} of Algorithm \ref{alg:2}, 
we sample a formula $f_{exp}$ from $\mathcal{G}$ as the formula to be explored in each search iteration, 
Following the idea of observing the confidence level of each formula and the formulas being close, prior study proposed an $Expectation$ value defined by Equation \ref{equ:exp} on each node to estimate how reliable the formula it represents can be for solving $q$ \cite{coling}. 
In each iteration of search, the node with the greatest $Expection$ value among unexplored nodes is selected as the node whose formula $f_{exp}$ is going to be explored.

\begin{align}
    n.Expectation = & \sum_{d=0}^{3} w_d * max\{n^*.score\ |\ n^* \in \mathcal{G}, distance(n,n^*)\leqslant d\} \label{equ:exp}\\
    w = &\ [0.4,\ 0.3,\ 0.2,\ 0.1] \label{equ:wexp}
\end{align}

\subsubsection{Formula Mutation}

As shown in line \ref{line:mutate} of Algorithm \ref{alg:2}, 
mutations from $f_{exp}$ are generated to expand the graph $\mathcal{G}$ at the end of each search iteration.
Here, we generate new formulas whose distance from $f_{exp}$ is one by inserting, removing, or modifying one single formula term. 
The freshly generated formulas are added to $\mathcal{G}$ if they do not exist in $\mathcal{G}$ yet, and the relevant edges should also be updated to keep $\mathcal{G}$ conforming to its definition.

\vspace{5pt}
After the search reaches the maximum number of iterations, the formula on which the highest accuracy is achieved is considered the optimal formula $f_{best}$ for solving $q$. 
$f_{best}$ is returned with the accuracy it achieves after \texttt{CleanUp} as the result of \texttt{Search}.
Together with \texttt{Reflect}, we proposed this \texttt{CleanUp} procedure to reduce the redundancy of the formulas obtained through a heuristic search.
The details of this \texttt{CleanUp} procedure are presented in Section \ref{sec:clean_reflect}.
If none of the formulas achieve a non-zero accuracy, \texttt{Search} returns \texttt{None}.

\subsection{Formula Clean-Up and Reflection} \label{sec:clean_reflect}

\setlength{\tabcolsep}{8pt}
\begin{table}[t]
\centering
\caption{Examples of the prolixity issue in formulas.} \label{tab:2}
\begin{tabular}{c l}
\toprule
$q$ & \texttt{Find the sum of $\langle$N0$\rangle$ and $\langle$N1$\rangle$.}\\
\midrule
$f_0$ & \texttt{[(+,\;N0,\;N1)]} \\
$f_1$ & \texttt{[(-,\;N0,\;N1),\;(+,\;N0,\;N1)]} \\
$f_2$ & \texttt{[(+,\;N0,\;N0),\;(+,\;M0,\;N1),\;(-,\;M1,\;N0)]} \\
\bottomrule
\end{tabular}
\end{table}

Through our study and experiment on mathematical problems with the learning method introduced above, a phenomenon we observed is that the formulas derived via a heuristic search occasionally exhibit unnecessary prolixity.
Table \ref{tab:2} illustrates this phenomenon on a simple problem. 
In contrast to $f_0$, which can be considered a valid and concise formula for solving the given $q$, 
even though $f_1$ expresses the same computation as $f_0$, its first term does not virtually contribute to the subsequent computations so that it becomes redundant.
This kind of redundant terms appears more or less in the search result because the search process advances by generating mutations via inserting, removing, and modifying formula terms.
$f_2$ illustrates another case of prolixity, whereas it is caused by unnecessary inverse operations instead of unused terms.
Including these prolix formulas into the training data of inference models without proper processing can adversely affect the learning efficiency and generalization capacity.
In view of this, we proposed the \texttt{CleanUp} and \texttt{Reflect} procedures in our study.

Concretely, with the first case of prolixity caused by unused terms named ``type-I prolixity'' and the second case caused by redundant computations named ``type-II prolixity'', \texttt{CleanUp} is utilized to eliminate the type-I prolixity, 
we first perform \texttt{CleanUp} in line \ref{line:cleanup} of Algorithm \ref{alg:2} to eliminate the type-I prolixity before a formula is returned as the search result.
For \texttt{CleanUp}, we simply verify whether each term $S_i$ is ever referred to as \texttt{Mi} in the subsequent terms, and remove the terms that are not. 
Note that the reference tokens referring to formula terms after the removed terms should be adjusted correspondingly to keep the reference intact.

As for the type-II prolixity, given that it is highly related to the particular properties of certain operations and thus may not be easily identified without injecting external knowledge, which is a practice we are committed to avoiding to uphold the universal applicability of the learning framework,
we propose to leverage the inference model to make this judgment.
As for the \texttt{Reflect} procedure performed at the end of each learning loop in line \ref{line:reflect} of Algorithm \ref{alg:1}, we let \texttt{PolicyNet} reperform the inference on each $q$ $\in$ $\mathbb{L}$. 
With $f_0$ and $accu_{f_0}$ denote the original formula and accuracy recorded for $q$ in $\mathbb{L}$, and $\hat{f}$ denotes the fresh inference of formula made by \texttt{PolicyNet}, we first examine the effectiveness of $\hat{f}$ with $\mathbb{I}$ on \{$num$\} and \{$a$\} to obtain its accuracy $accu_{\hat{f}}$.
Then, if $accu_{\hat{f}}$ $\geqslant$ $accu_{f_0}$ and the length (i.e., the number of terms) of $\hat{f}$ is not longer than $f_0$, we update the formula and accuracy recorded for $q$ in $\mathbb{L}$ with ($\hat{f}$, $accu_{\hat{f}}$).
Intuitively, in this case, we consider that the inference model suggests an alternative formula that also effectively solves the relevant problem while consuming fewer calculation terms or exhibiting higher consistency with existing formulas.
As a result, the previously recorded formula can be reasonably updated.
This mechanism also contributes to the exclusion of formulas whose effectiveness is misjudged.
We studied the effect of this reflection mechanism with an ablation study and reported the result in Section \ref{sec:abr}.

\section{Asynchrony and Parallelization in Search Process} \label{sec:async_paral}

\begin{figure}[t]
\centering
\begin{minipage}{0.9\textwidth}
\begin{algorithm}[H]
\caption{Asynchronous Formula Scoring} \label{alg:3}
\begin{algorithmic}[1]
\STATE \textbf{func} \texttt{Score}($q$,\ $f$,\ \{$num$\},\ \{$a$\},\ \texttt{pipe})
    \STATE \hspace{1em} \texttt{pipe}.send($q$, $f$)
    \STATE \hspace{1em} $f.eval$\_$score$ $\gets$ Accuracy($\mathbb{I}(f,\ \{num\})$,\ \{$a$\}) \hspace{1em} \hfill $\triangleright$ High workload on CPU
    \STATE \hspace{1em} $f.pri$\_$score$ $\gets$ \texttt{pipe}.receive()
    \STATE \hspace{1em} \textbf{return} $f.pri$\_$score + \beta * f.eval$\_$score$
\item[]
\STATE \textbf{func} \texttt{GPU\_Handler}(\texttt{PolicyNet},\ \texttt{pipe})
    \STATE \hspace{1em} \textbf{while True do}
        \STATE \hspace{2em} $q$, $f$ $\gets$ \texttt{pipe}.receive()
        \STATE \hspace{2em} $f.pri$\_$score$ $\gets$ $\hat{p}(f|q, \theta_{P})$ \hfill $\triangleright$ High workload on GPU
        \STATE \hspace{2em} \texttt{pipe}.send($f.pri$\_$score$)
    \STATE \hspace{1em} \textbf{end while}
\end{algorithmic}
\end{algorithm}
\end{minipage}
\end{figure}

In addition to the techniques introduced above, we also incorporate the asynchrony and parallelization practices in the search process to further improve the search efficiency with engineering efforts.
Here, our proposal is derived from a fact we observed in the search process that the workloads on CPU and GPU occur alternately.
Specifically, the computational load on CPU primarily arises during the update of the formula graph and the evaluation of formulas. 
On the other hand, the computational load on GPU primarily arises during the scoring of formulas and the calculation of $p(f|q,\theta_{P})$ given \texttt{PolicyNet}.
In consideration that these computations also occur sequentially in the original search process, the overall usage of both CPU and GPU ends up being low.
This leaves us room for improvement by employing asynchrony and parallelization practices.

To make full use of the computational resources on both CPU and GPU, we first reimplement the formula scoring in an asynchronous manner as presented in Algorithm \ref{alg:3}.
Here, the \texttt{Score} function, which is a subprocedure of the main search process, and the \texttt{GPU\_Handler} function are executed in two separate processes.
They communicate with a data structure \texttt{pipe}.
Generally, after receiving the formula $f$ to be scored and the associated ($q$,\ \{$num$\},\ \{$a$\}), the formula scoring procedure first delivers the task of calculating the likelihood of $f$ with inference model to the \texttt{GPU\_Handler}. 
Then, the formula evaluation is performed during the time spent waiting for the calculation on \texttt{GPU\_Handler} to be done.
The asynchronous execution of these two tasks enables the concurrent usage of computational resources on both CPU and GPU.

Furthermore, we also implement a more general parallelized learning framework for the entire learning process.
In this framework, the \texttt{Search} procedure in line \ref{line:alg_search} of Algorithm \ref{alg:1} is parallelized into multiple search processes, which allows multiple search attempts on different learning instances to be performed concurrently.
In addition, there can also be multiple GPU Handlers to maximize resource utilization in a multi-GPU environment and serve more search processes. 
To achieve this, a central controller is responsible for not only organizing all the other procedures presented in Algorithm \ref{alg:1} but also the synchronization of inference models across multiple GPU Handlers.
Concretely, after all the search processes finish and the \texttt{PolicyNet} is optimized as shown by line \ref{line:train}, the central controller broadcasts the optimized parameters of \texttt{PolicyNet} to all the GPU Handlers to keep inference models synchronized.

\section{Experiments}

In this study, we assesses the effectiveness of our methods on the Mathematics dataset \cite{math_dm}. We chose the Mathematics dataset for our experiments for several reasons.
Firstly, the Mathematics dataset offers mathematical problems of diverse difficulties across multiple domains and problem categories, which allows us to evaluate our approach in various scenarios.
Secondly, although neuro-symbolic methods are highly suitable for solving the problems in the Mathematics dataset, the original training data provided by the Mathematics dataset only comprises the question texts and final answers.
This allows us to verify our weakly-supervised approach for learning intermediate labels in neuro-symbolic methods and make fair comparisons with other methods.
Lastly, the Mathematics dataset affords millions of pre-generated question--answer samples for each problem category and supports unlimited data generation.
This ensures a sufficient amount of training data for organizing and verifying the learning of formulas from scratch.

\subsection{Experimental Setup}

In consideration of the data types and computations that can be supported, as well as the requirements for formula examination, we collected a subset from the Mathematics dataset that satisfies two criteria.
First, the answers to questions should be a single rational number.
Second, each template should correspond to a minimum of three raw questions on average after data preprocessing. 
For the inference model, We employ the Memory-Interactive Learning Engine (MILE) \cite{mile} with an LSTM \cite{lstm} encoder as the \texttt{PolicyNet} because its architecture is naturally compatible with the structure of the DSL we adopt.
For the learning process, following the conventional practice adopted by existing research, we conducted the training and evaluation on each problem category separately.
As for the hyperparameters, we let $max\_loop$ $=$ 100000, $max\_iter$ $=$ 10000, $\beta$ $=$ 1.0, the number for initial nodes in formula graph $N$ $=$ $10$ and $M$ $=$ $10$, and the acceptable threshold for question answering accuracy $\gamma$ $=$ $0.99$.
The set of operations we utilized is shown in Appendix \ref{apd:1} and \ref{apd:2}.

\subsection{Primary Result}

\setlength{\tabcolsep}{5.2pt}
\begin{table}[t]
\centering
\caption{Primary results on a subset of the Mathematics dataset.} \label{tab:exp1}
\begin{tabular}{r | c c | c c c c c c c c | c c c}
\toprule
    & \multicolumn{2}{c|}{Algebra} & \multicolumn{8}{c|}{Arithmetic} & \multicolumn{3}{c}{Measurement}\\
    Category & L1 & SN & AS & ASB & ASM & Di & Mi & Mu & MD & NI & Co & Co* & Ti \\
    \midrule
    \texttt{\#}Train & 1491 & 18 & 18 & 2 & 15203 & 4 & 36091 & 8 & 4807 & 11 & 7267 & 1163 & 8 \\
    \texttt{\#}Solved & 1474 & 18 & 18 & 2 & 15202 & 4 & 30317 & 8 & 4803 & 11 & 7151 & 1146 & 8 \\
    \% & 98.8 & 100 & 100 & 100 & 99.9 & 100 & 84.0 & 100 & 99.9 & 100 & 98.4 & 98.5 & 100 \\
    \midrule
    Test & 95.2 & \textbf{100} & \textbf{100} & \textbf{100} & \textbf{99.8} & \textbf{100} & 
    \textbf{73.5} & \textbf{100} & \textbf{99.7} & \textbf{99.8} & 65.8 & \textbf{94.5} & \textbf{100} \\
\midrule
    \multicolumn{14}{c}{End-to-end solutions} \\
\midrule
    Transformer\textsuperscript{\textdagger} & \textbf{98.6} & 92.3 & 99.6 & 98.9 & 98.2 & 87.7 & 68.3 & 74.5 & 95.0 & 92.7 & \multicolumn{2}{c}{92.9} & 99.4 \\
    LSTM\textsuperscript{\textdagger}  & 84.1 & 50.1 & 95.0 & 92.6 & 95.4 & 82.2 & 61.5 & 51.5 & 95.9 & 89.3 & \multicolumn{2}{c}{78.1} & 72.4 \\
\midrule
    \multicolumn{14}{c}{LLM-based solutions} \\
\midrule
    GPT-3.5\,/\,zs & 7.2 & 31.0 & 92.1 & 33.4 & 25.5 & 72.8 & 4.1 & 58.4 & 2.7 & 6.8 & \multicolumn{2}{c}{62.6} & 63.9 \\
    GPT-3.5\,/\,fs & 9.7 & 43.6 & 91.0 & 41.7 & 21.7 & 75.5 & 3.9 & 60.9 & 2.9 & 7.6 & \multicolumn{2}{c}{66.6} & 64.0 \\         
\midrule
    \multirow{3}{*}{GPT-4 CoT} & 97.3 & 89.8 & 96.5 & 12.2 & 98.9 & 80.7 & 67.2 & 82.4 & 73.1 & 44.0 & \multicolumn{2}{c}{93.7} & 94.6 \\
    & 97.0 & 87.8 & 95.2 & 9.4 & 98.7 & 79.3 & 66.3 & 80.8 & 70.1 & 35.0 & \multicolumn{2}{c}{92.1} & 93.9 \\
    & 97.2 & 89.6 & 96.1 & 12.4 & 99.2 & 81.6 & 66.8 & 82.6 & 73.4 & 43.6 & \multicolumn{2}{c}{92.7} & 94.8 \\
\bottomrule
\multicolumn{14}{c}{}\\[-7pt]
\multicolumn{14}{l}{\footnotesize L1: linear\_1d \; SN: sequence\_next\_term \; AS: add\_or\_sub \; ASB: add\_or\_sub\_in\_base \; ASM: add\_sub\_multiple \; Di: div}\\
\multicolumn{14}{l}{\footnotesize Mi: mixed \; Mu: mul \; MD: mul\_div\_multiple \; NI: nearest\_integer\_root \; Co: conversion \; Ti: time} \\[5pt]
\multicolumn{14}{l}{\textsuperscript{\textdagger} baselines provided by \cite{math_dm}}
\end{tabular}
\end{table}

The primary results of our experiments are presented in Table \ref{tab:exp1}.
In this table, in the first three rows, we first present the number of unique question templates obtained in the training set of each problem category after data preprocessing and the number and proportion of templates solved by the formula exploration in the training stage.
Generally, these results indicate the effectiveness of the formula exploration and weakly-supervised formula learning. 
In the following rows, we show the test accuracy achieved by our methods and baselines on the test sets. 
Here, ``Transformer" and ``LSTM" are the baselines provided by \cite{math_dm} that implement end-to-end training. 
Moreover, we also investigated the performance of modern LLMs.
Specifically, we first employ the GPT-3.5 model \cite{gpt3.5} to perform zero-shot and few-shot in-context learning and report the result as ``GPT-3.5 / zs" and ``GPT-3.5 / fs". 
The prompts we use for GPT-3.5 are shown in Appendix \ref{apd:3}.
On the other hand, we leverage the GPT-4 model \cite{gpt4} to perform the Chain-of-Thought reasoning by utilizing the three prompts suggested by \cite{zsr} that resulted in the best performance.
The details of these prompts are shown in Appendix \ref{apd:4}.
For generating the response with the GPT-3.5 and GPT-4 models, we set the $temperature$ to 0.5 and $top$-$p$ to 1.0.

Based on these results, it can first be concluded that the weakly-supervised formula learning is capable of acquiring valid formulas in most problem categories. 
The proportion of question templates that are solved drops to some degree in some challenging problem categories represented by ``mixed", which contains problems involving nested and mixed arithmetic calculations and results in an enormous space of possible formulas.
However, the accuracy achieved on the test set is still desirable compared to the baselines performing end-to-end training and prompting LLMs to acquire the answer.

As for the results on other problem categories, it can be noticed that on ``linear\_1d", 
the test accuracy of our neuro-symbolic practice underperforms end-to-end methods.
We attribute this failure to the limitation exposed in generalization. 
In this problem category, although the proportion of question templates solved in the training stage remains high, there is a gap between this proportion and the final accuracy achieved on the test set, which suggests the generalization of formulas from the training set to the test set on such a problem category can be challenging. 
A similar phenomenon can also be observed in the problem category ``conversion".
For this problem category, we provide an alternative experimental setting ``Co*" where we replace numerals in English with their corresponding numerical digits (e.g., replacing the word ``twelve" in question texts with ``12"), which can be identified as number tokens during the data preprocessing. 
This allows more combinations for question templates and results in fewer unique templates.
Regarding the result, although a similar proportion of question templates are solved in ``Co" and ``Co*", ``Co*" achieves much higher test accuracy thanks to the reduced difficulty in generalization.
As a feasible solution to address the generalization problem, given that we are now implementing MILE with a simple LSTM encoder in consideration of the computational cost in formula search, 
we suggest a retraining for the inference model with more powerful encoders utilizing the acquired pairs of ($q$, $f$) after the weakly-supervised learning stage is completed.
This practice can also facilitate the balance between reducing the computational cost spent on the inference model in the weakly-supervised learning stage and enhancing the performance in the stage of inference.

\subsection{Ablation Study}

\subsubsection{Effectiveness of Formula Graph}

\begin{figure}[t]
\centering
\includegraphics[width=\textwidth]{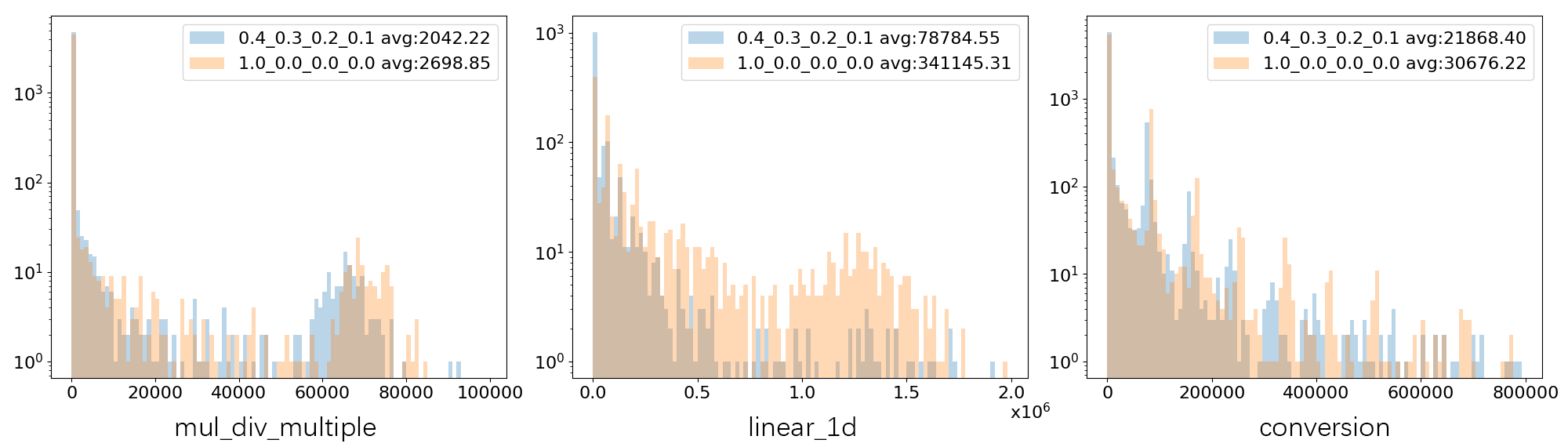}
\caption{The histogram that shows the number of search iterations consumed to acquire each valid formula on three representative problem categories: mul\_div\_multiple, linear\_1d, and conversion.}
\label{fig:ab1}
\end{figure}

In addition to the primary experiment, we also conducted ablation studies to verify the effectiveness of several crucial proposals and mechanisms employed in our learning framework. 
The first factor we investigated here is the effectiveness of the graph-based formula exploration technique, which is a fundamental proposal made in our work.
To perform an ablation study, we set the weight $w$ for summarizing formula score in Equation \ref{equ:wexp} 
from its default scale $[0.4,\ 0.3,\ 0.2,\ 0.1]$ to $[1.0,\ 0.0,\ 0.0,\ 0.0]$.
This prevents the formula sampling in line \ref{line:sample} of Algorithm \ref{alg:2}
from observing the neighbors of each formula and makes the graph structure degenerate into an ordinary collection of independent formula nodes.
In this experiment, we investigated the number of search iterations consumed in finding valid formulas for each learning sample with regular and degenerate $w$ 
on three representative problem categories: mul\_div\_multiple, linear\_1d, and conversion.
The results are presented with histograms in Figure \ref{fig:ab1}.
From this result, it can be noticed that the search procedure with a regular $w$ generally consumed fewer search iterations to discover each valid formula.
This can be observed both from the distribution presented by the histogram and the average number of consumed search iterations.
This result demonstrates that the efficiency of the formula search is improved by the search mechanism organized on the formula graph.

\subsubsection{Effectiveness of Reflection Mechanism} \label{sec:abr}

\setlength{\tabcolsep}{6pt}
\begin{table}[t]
\centering
\begin{minipage}{0.9\textwidth}
\caption{Test accuracy achieved without and with the formula reflection on several challenging problem categories.} \label{tab:ab2}
\centering
\begin{tabular}{r | c | c c c | c }
\toprule
    & Algebra & \multicolumn{3}{c|}{Arithmetic} & Measurement\\
    Category & L1 & ASM & Mi & MD & Co \\
\midrule
    w/o reflection & 93.3 & 98.5 & 65.4 & 98.8 & 61.0 \\
    w/ reflection & 95.2 & 99.8 & 73.5 & 99.7 & 65.8 \\
\bottomrule
\multicolumn{6}{c}{}\\[-7pt]
\multicolumn{6}{l}{\footnotesize L1: linear\_1d \; ASM: add\_sub\_multiple \; Mi: mixed} \\
\multicolumn{6}{l}{\footnotesize MD: mul\_div\_multiple \; Co: conversion} \\
\end{tabular}
\end{minipage}
\end{table}

Furthermore, we investigated the effect of the reflection mechanism we proposed in Section \ref{sec:clean_reflect}.
In this experiment, we simply activated and deactivated the \texttt{Reflect} procedure adopted in line \ref{line:reflect} of Algorithm \ref{alg:1}.
This experiment is conducted on several relatively challenging problem categories with diverse question templates, 
and the result is presented in Table \ref{tab:ab2},
Based on these results, it can be concluded that the reflection mechanism generally enhances the generalization capacity of inference thanks to the mitigation of the prolixity issue and the improvement in the consistency of formulas.

\subsubsection{Asynchrony and Parallelization}

\begin{table}[t]
\centering
\begin{minipage}{0.9\textwidth}
\caption{Average number of formulas that are examined per second in the formula exploration process. ``Legacy" refers to the classic implementation without any asynchrony and parallelization practice.}  \label{tab:ab3}
\centering
\begin{tabular}{r c}
\toprule
    & \texttt{\#}formula / sec \\
\midrule
    Legacy & 102 \\
    2 GPUs, 4 search processes & 2881 \\
    4 GPUs, 8 search processes & 4558 \\
    4 GPUs, 16 search processes & 4796 \\
\bottomrule
\end{tabular}
\end{minipage}
\end{table}

To evaluate the improvement in the efficiency of formula searches resulting from the asynchrony and parallelization practices we introduced in Section \ref{sec:async_paral}, we measured the rate of formula examination achieved under difference hardware expenses.
Here, the experiment is conducted on a computing server operating with Ubuntu 18.04 and Python 3.10.5.
The model of the CPU is 32-Core AMD EPYC 7452, and the model of the GPU is 40GB NVIDIA A100.
The results are reported in Table \ref{tab:ab3}.
In these results, the formula examining rate is averaged across three problem categories: mul\_div\_multiple, linear\_1d, and conversion.
``Legacy" refers to the classic implementation without any asynchrony and parallelization practice.

From these results, it can first be concluded that the asynchrony and parallelization practices we introduced in Section \ref{sec:async_paral} largely accelerate the formula examination.
Moreover, we can observe that the formula examining rate is majorly influenced by the number of GPUs, which virtually affects the progress rate of search iterations as each search iteration must wait for the computation on GPUs to be finished to obtain the scoring results for the new formulas.
When the GPUs have reached their maximum workload, further increasing the number of search processes contributes marginally to the overall exploration efficiency.
Accordingly, we suggest prioritizing fulfilling the computing power requirements on GPUs to maximize the efficiency of formula exploration.

\subsubsection{Comparison with Previous Work}

\begin{figure}[t]
\centering
\includegraphics[width=\textwidth]{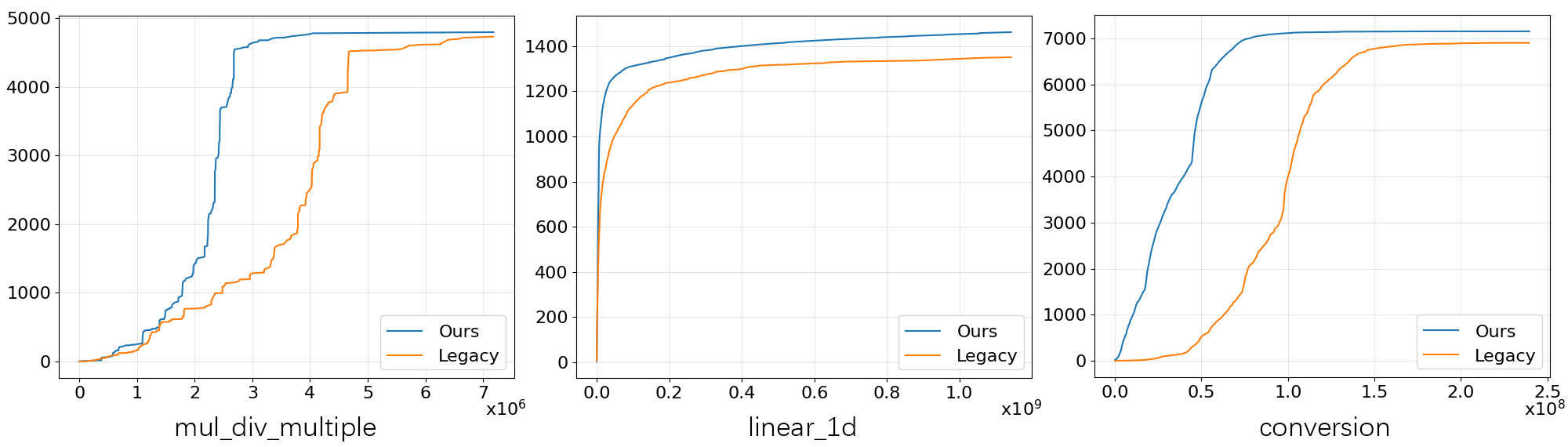}
\caption{The progress of the formula search illustrated by the correlation between the total number of search iterations and the counts of solved problems on three representative problem categories. ``legacy'' denotes the implementation of \cite{coling}.} \label{fig:ab2}
\end{figure}

\setlength{\tabcolsep}{9pt}
\begin{table}[t]
\centering
\begin{minipage}{0.9\textwidth}
\caption{Test accuracy achieved by our method and existing work on three representative problem categories.} \label{tab:ab4}
\centering
\begin{tabular}{r | c | c | c }
\toprule
    Category & L1 & Mi & Co \\
\midrule
    Ours & 95.2 & 73.5  & 65.8 \\
    Legacy \cite{coling} & 75.0 & 58.9 & 58.4 \\
\bottomrule
\multicolumn{4}{c}{}\\[-7pt]
\multicolumn{4}{l}{\footnotesize L1: linear\_1d \; Mi: mixed \; Co: conversion} \\
\end{tabular}
\end{minipage}
\end{table}

Finally, we performed a direct comparison between our proposals and the existing formula exploration approach \cite{coling} with a quantitative metric to assess the test accuracy as well as a qualitative metric to illustrate the formula search progress during the learning stage.
The results are shown in Table \ref{tab:ab4} and Figure \ref{fig:ab2}, respectively.
Both these results demonstrate that our refined learning methods achieve higher formula exploration efficiency and better test performance compared to the classic learning approach.

\section{Further Discussion}

In light of the rapid advancements in the techniques and applications of LLMs nowadays, we consistently remain concerned with the correlation between our proposals and LLMs.
Generally, we consider that the major contribution of our proposals does not conflict with the application of LLMs, and the reasons are as follows.
Firstly, as we mentioned at the beginning, even though LLMs exhibit a rich practicality in generating symbolic instructions for neuro-symbolic reasoning, the reliability of these instructions is not always assured.
Accordingly, in cases that the prediction of LLMs involves false symbolic instructions, our proposals can act in eliminating these noises.
Secondly, our methods can also work in conjunction with LLMs through employing LLMs as our \texttt{PolicyNet} to suggest reasonable prior formula inferences for the initialization of our formula graph, or utilizing LLMs to estimate the likelihood of formulas for the formula scoring.
The search process can also benefit from the extensive prior knowledge embedded in pretrained LLMs.

As for the potential limitation of our work, the major concern we retain still lies in the feasibility of weak-supervised learning in extremely large search spaces. 
This limitation arises due to two primary facts: the size of the search space increases exponentially with the growth of the count of necessary calculations required for solving associated problems, and the learning approaches we studied in this work majorly focus on the problem setting that no prior knowledge is provided, which requires the learning process to start completely from scratch.
In practice, we also observe that our learning practice rarely succeeds in discovering valid formulas exceeding ten operations in their length.
With respect to this limitation, we consider endowing the learning procedure with appropriate prior knowledge to be a feasible solution, which also serves as a significant direction for future work.

\section{Conclusion}

This paper discussed the application and limitation of neuro-symbolic approaches in mathematical reasoning tasks.
In these approaches, intermediate symbolic labels, named formulas, are indispensable for the training process but can be hard to acquire.
Moreover, existing weakly-supervised learning techniques proposed by previous work can be inadequate to solve more complex and challenging mathematical problems, and their compatibility with computations in more arbitrary and non-tree-structured forms is also questionable.
To address these problems, in this work, we developed an advanced learning framework on the basis of prior work to facilitate the weakly-supervised learning of formulas given the problem inputs and final answers.
In this framework, adhering to the principle established by prior work that organizes recurrent exploration process and heuristic search, we extend the formulas to a more generalized symbolic notation system that support functional DSLs, and further enhance the efficiency of formula exploration with our multiple novel proposals.
Our experiments on the Mathematics dataset illustrates the efficacy of neuro-symbolic approaches powered by our weakly-supervised formula exploration.
Furthermore, the ablation study confirms the effectiveness of the novel practices and mechanisms integrated into the learning framework.
In view of this, we consider that our proposals offer a valid and advanced approach to the application of neuro-symbolic practices in mathematical reasoning tasks.

\bigskip

\bibliographystyle{unsrt}  
\bibliography{references}  

\begin{thebibliography}{10}

\bibitem{ns0}
Warren~S McCulloch and Walter Pitts.
\newblock A logical calculus of the ideas immanent in nervous activity.
\newblock {\em The bulletin of mathematical biophysics}, 5:115--133, 1943.

\bibitem{ns1}
Artur~d’Avila Garcez, Sebastian Bader, Howard Bowman, Luis~C Lamb, Leo de~Penning, BV~Illuminoo, and Hoifung Poon.
\newblock Neural-symbolic learning and reasoning: A survey and interpretation.
\newblock {\em Neuro-Symbolic Artificial Intelligence: The State of the Art}, 342(1):327, 2022.

\bibitem{ns2}
Md~Kamruzzaman Sarker, Lu~Zhou, Aaron Eberhart, and Pascal Hitzler.
\newblock Neuro-symbolic artificial intelligence: Current trends.
\newblock {\em AI Communications}, 34, 2022 2022.

\bibitem{math23}
Yan Wang, Xiaojiang Liu, and Shuming Shi.
\newblock Deep neural solver for math word problems.
\newblock In {\em Proceedings of the 2017 Conference on Empirical Methods in Natural Language Processing}, pages 845--854, 2017.

\bibitem{mathqa}
Aida Amini, Saadia Gabriel, Shanchuan Lin, Rik Koncel-Kedziorski, Yejin Choi, and Hannaneh Hajishirzi.
\newblock {M}ath{QA}: Towards interpretable math word problem solving with operation-based formalisms.
\newblock In {\em Proceedings of the 2019 Conference of the North {A}merican Chapter of the Association for Computational Linguistics: Human Language Technologies, Volume 1 (Long and Short Papers)}, pages 2357--2367, 2019.

\bibitem{gsm8k}
Karl Cobbe, Vineet Kosaraju, Mohammad Bavarian, Jacob Hilton, Reiichiro Nakano, Christopher Hesse, and John Schulman.
\newblock Training verifiers to solve math word problems.
\newblock {\em arXiv preprint arXiv:2110.14168}, 2021.

\bibitem{mawps}
Rik Koncel-Kedziorski, Subhro Roy, Aida Amini, Nate Kushman, and Hannaneh Hajishirzi.
\newblock {MAWPS}: A math word problem repository.
\newblock In {\em Proceedings of the 2016 Conference of the North {A}merican Chapter of the Association for Computational Linguistics: Human Language Technologies}, pages 1152--1157, San Diego, California, June 2016.

\bibitem{cot}
Jason Wei, Xuezhi Wang, Dale Schuurmans, Maarten Bosma, Ed~Chi, Quoc Le, and Denny Zhou.
\newblock Chain of thought prompting elicits reasoning in large language models.
\newblock {\em arXiv preprint arXiv:2201.11903}, 2022.

\bibitem{math_fix}
Yining Hong, Qing Li, Daniel Ciao, Siyuan Huang, and Song-Chun Zhu.
\newblock Learning by fixing: Solving math word problems with weak supervision.
\newblock {\em Proceedings of the AAAI Conference on Artificial Intelligence}, 35(6):4959--4967, 2021.

\bibitem{coling}
Yuxuan Wu and Hideki Nakayama.
\newblock Weakly supervised formula learner for solving mathematical problems.
\newblock In {\em Proceedings of the 29th International Conference on Computational Linguistics}, pages 1743--1752, 2022.

\bibitem{math_dm}
David Saxton, Edward Grefenstette, Felix Hill, and Pushmeet Kohli.
\newblock Analysing mathematical reasoning abilities of neural models.
\newblock In {\em International Conference on Learning Representations}, 2019.

\bibitem{math_g0}
Jiaqi Chen, Jianheng Tang, Jinghui Qin, Xiaodan Liang, Lingbo Liu, Eric Xing, and Liang Lin.
\newblock Geoqa: A geometric question answering benchmark towards multimodal numerical reasoning.
\newblock In {\em Findings of the Association for Computational Linguistics: ACL-IJCNLP 2021}, pages 513--523, 2021.

\bibitem{math_atp}
Kunhao Zheng, Jesse~Michael Han, and Stanislas Polu.
\newblock minif2f: a cross-system benchmark for formal olympiad-level mathematics.
\newblock In {\em International Conference on Learning Representations}, 2021.

\bibitem{ldj}
Kaiyu Yang, Aidan~M Swope, Alex Gu, Rahul Chalamala, Peiyang Song, Shixing Yu, Saad Godil, Ryan Prenger, and Anima Anandkumar.
\newblock Leandojo: Theorem proving with retrieval-augmented language models.
\newblock In {\em Thirty-seventh Conference on Neural Information Processing Systems Datasets and Benchmarks Track}, 2023.

\bibitem{mmlu}
Dan Hendrycks, Collin Burns, Steven Basart, Andy Zou, Mantas Mazeika, Dawn Song, and Jacob Steinhardt.
\newblock Measuring massive multitask language understanding.
\newblock In {\em International Conference on Learning Representations}, 2020.

\bibitem{palm}
Aakanksha Chowdhery, Sharan Narang, Jacob Devlin, Maarten Bosma, Gaurav Mishra, Adam Roberts, Paul Barham, Hyung~Won Chung, Charles Sutton, Sebastian Gehrmann, et~al.
\newblock Palm: Scaling language modeling with pathways.
\newblock {\em arXiv preprint arXiv:2204.02311}, 2022.

\bibitem{zs1}
Vered Shwartz, Peter West, Ronan Le~Bras, Chandra Bhagavatula, and Yejin Choi.
\newblock Unsupervised commonsense question answering with self-talk.
\newblock In {\em Proceedings of the 2020 Conference on Empirical Methods in Natural Language Processing (EMNLP)}, pages 4615--4629, 2020.

\bibitem{zs2}
Laria Reynolds and Kyle McDonell.
\newblock Prompt programming for large language models: Beyond the few-shot paradigm.
\newblock In {\em Extended Abstracts of the 2021 CHI Conference on Human Factors in Computing Systems}, pages 1--7, 2021.

\bibitem{zsr}
Takeshi Kojima, Shixiang Gu, Machel Reid, Yutaka Matsuo, and Yusuke Iwasawa.
\newblock Large language models are zero-shot reasoners.
\newblock In {\em Advances in Neural Information Processing Systems}, volume~35, pages 22199--22213, 2022.

\bibitem{wolfram}
Stephen Wolfram.
\newblock Chatgpt gets its ``wolfram superpowers"!.
\newblock \url{https://writings.stephenwolfram.com/2023/03/chatgpt-gets-its-wolfram-superpowers}, 2023.

\bibitem{mile}
Yuxuan Wu and Hideki Nakayama.
\newblock {MILE}: Memory-interactive learning engine for neuro-symbolic solutions to mathematical problems, 2024.

\bibitem{lstm}
Sepp Hochreiter and J{\"u}rgen Schmidhuber.
\newblock Long short-term memory.
\newblock {\em Neural computation}, 9(8):1735--1780, 1997.

\bibitem{gpt3.5}
{OpenAI}.
\newblock {GPT}-3.5 \url{[ gpt-3.5-turbo-0125 ]}.
\newblock \url{https://www.openai.com}, 2024.

\bibitem{gpt4}
{OpenAI}.
\newblock {GPT}-4 \url{[ gpt-4o-2024-05-13 ]}.
\newblock \url{https://www.openai.com}, 2024.

\end{thebibliography}

\clearpage

\appendix

\section{Definition of mathematical operators} \label{apd:1}

\setlength{\tabcolsep}{8pt}
\begin{table}[H]
\centering
\caption{The set of operators we adopted in our experiments and their functionality.}
\begin{tabularx}{0.9\textwidth}{r X}
    \toprule
        \texttt{$+$} & \texttt{(+,\;a,\;b)} returns the value $(a+b)$.\\
        \midrule
        \texttt{$-$} & \texttt{(-,\;a,\;b)} returns the value $(a-b)$. \\
        \midrule
        \texttt{$\times$} & \texttt{($\times$,\;a,\;b)} returns the value $(a \times b)$. \\
        \midrule
        \texttt{$\div$} & \texttt{($\div$,\;a,\;b)} returns the value $(a \div b)$. \\
    \midrule
        $\hat{}$\; & \texttt{(\^{},\;a,\;b)} returns the value $a$ to the power of $b$. 
            It raises an error if $a$ is not positive and $b$ is not an integer. \\
        \midrule
        \texttt{ABS} & \texttt{(ABS,\;a)} returns the absolute value of $a$. \\
        \midrule
        \texttt{ROUND} & \texttt{(ROUND,\;a)} rounds $a$ up or down to the nearest integer. \\
        \midrule
        \texttt{BASE\_FROM} & \texttt{(BASE\_FROM,\;a,\;b)} converts $a$ from base $b$ to base 10. \\
        \midrule
        \texttt{BASE\_TO} & \texttt{(BASE\_TO,\;a,\;b)} converts $a$ from base 10 to base $b$. \\
    \bottomrule
\end{tabularx}
\end{table}

\section{Operators used in each problem category} \label{apd:2}

\begin{table}[H]
\centering
\caption{The sets of operators required by each problem category.}
\begin{tabular}{r l}
    \toprule
        \multicolumn{2}{c}{algebra}\\
        \midrule
        linear\_1d & $+$, $-$, $\times$, $\div$, \;$\hat{}$\; \\
        sequence\_next\_term & $+$, $-$, $\times$, $\div$ \\
    \midrule
        \multicolumn{2}{c}{arithmetic}\\
    \midrule
        add\_or\_sub & $+$, $-$, \texttt{ABS} \\
        add\_or\_sub\_in\_base & $+$, $-$, \texttt{BASE\_FROM}, \texttt{BASE\_TO} \\
        add\_sub\_multiple & $+$, $-$, \texttt{ABS} \\
        div & $\times$, $\div$ \\
        mixed & $+$, $-$, $\times$, $\div$ \\
        mul & $\times$, $\div$ \\
        mul\_div\_multiple & $\times$, $\div$ \\
        nearest\_integer\_root & $\times$, $\div$, \;$\hat{}$\; , \texttt{ROUND} \\
    \midrule
        \multicolumn{2}{c}{measurement}\\
    \midrule
        conversion & $+$, $-$, $\times$, $\div$ \\
        time & $+$, $-$, $\times$, $\div$, \texttt{ABS} \\       
    \bottomrule    
\end{tabular}
\end{table}

\clearpage

\section{Prompts used for GPT-3.5} \label{apd:3}

\begin{table}[H]
\centering
\begin{minipage}{0.9\textwidth}
\caption{The templates utilized for zero-shot and few-shot prompting in GPT-3.5. In few-shot in-context prompting, the example questions and answers are randomly sampled from the training set of the same problem category.}    
\begin{tabularx}{\linewidth}{r X}
\toprule
    zero-shot & Please answer the following question. You only need to give the final answer. \\
    & Q: \texttt{(question text)}\\
    & A: \\
\midrule
    few-shot & Please answer the question referring to the following examples. You only need to give the final answer. \\
    & Q: \texttt{(example question text \#1)}\\
    & A: \texttt{(example answer \#1)}\\
    & ... \\
    & Q: \texttt{(example question text \#6)}\\
    & A: \texttt{(example answer \#6)}\\
    & Q: \texttt{(question text)}\\
    & A: \\
\bottomrule    
\end{tabularx}
\end{minipage}
\end{table}

\section{Prompts used for GPT-4} \label{apd:4}

\begin{table}[H]
\centering
\begin{minipage}{0.9\textwidth}
\caption{The templates utilized on GPT-4 to perform Chain-of-Thought reasoning based on three prompts suggested by \cite{zsr} with the best performance.}
\begin{tabularx}{\linewidth}{r X}
\toprule
    CoT prompt \#1 & Q: \texttt{(question text)}\\
    & A: Let's think step by step.\\
\midrule
    CoT prompt \#2 & Q: \texttt{(question text)}\\
    & A: First,\\
\midrule
    CoT prompt \#3 & Q: \texttt{(question text)}\\
    & A: Let's think about this logically.\\
\bottomrule    
\end{tabularx}
\end{minipage}
\end{table}

\clearpage

\end{document}